\documentclass[10pt,twocolumn,letterpaper]{article}
\pdfoutput=1
\usepackage[pagenumbers]{cvpr} 

\usepackage{graphicx}
\usepackage{amsmath}
\usepackage{amssymb}
\usepackage{booktabs}
\usepackage[dvipsnames,table,xcdraw]{xcolor}

\usepackage{arydshln}
\usepackage{enumitem}
\usepackage{balance}
\usepackage{combelow}
\usepackage{tabularx}
\makeatletter
\def\adl@drawiv#1#2#3{%
        \hskip.5\tabcolsep
        \xleaders#3{#2.5\@tempdimb #1{1}#2.5\@tempdimb}%
                #2\z@ plus1fil minus1fil\relax
        \hskip.5\tabcolsep}
\newcommand{\cdashlinelr}[1]{%
  \noalign{\vskip\aboverulesep
           \global\let\@dashdrawstore\adl@draw
           \global\let\adl@draw\adl@drawiv}
  \cdashline{#1}
  \noalign{\global\let\adl@draw\@dashdrawstore
           \vskip\belowrulesep}}
\makeatother

\makeatletter
\newcommand{\dashrule}[1][black]{%
  \color{#1}\rule[\dimexpr.5ex-.2pt]{4pt}{.4pt}\xleaders\hbox{\rule{4pt}{0pt}\rule[\dimexpr.5ex-.2pt]{4pt}{.4pt}}\hfill\kern0pt%
}
\newcommand{\rulecolor}[1]{%
  \def\CT@arc@{\color{#1}}%
}
\makeatother

\makeatletter
\DeclareRobustCommand\onedot{\futurelet\@let@token\@onedot}
\def\@onedot{\ifx\@let@token.\else.\null\fi\xspace}

\def\eg{\emph{e.g}\onedot} 
\def\ie{\emph{i.e}\onedot} 
 
 \def\vs{\emph{vs}\onedot}
 
\def\etal{\emph{et al}\onedot}
\makeatother

\usepackage{wrapfig}
\usepackage[stable]{footmisc}

\usepackage{multirow}
\usepackage{makecell}
\usepackage{mathtools}
\usepackage{algorithm}
\usepackage{algpseudocode}
\usepackage{algorithmicx}
\usepackage{eqparbox,array}
\usepackage{pifont}
\usepackage{sidecap}
\usepackage{caption}
\usepackage{xspace}

\newcommand*\numcircledmod[1]{\raisebox{.5pt}{\textcircled{\raisebox{-.9pt} {#1}}}}

\usepackage{tabulary}

\definecolor{Orange}{rgb}{1,0.5,0.0}
\definecolor{azure}{rgb}{0.0, 0.5, 1.0}
\newcommand{\cmark}{\ding{51}}%
\newcommand{\xmark}{\ding{55}}%

\newcommand{\method}{BURN\xspace}

\usepackage[pagebackref,breaklinks,colorlinks]{hyperref}

\usepackage[capitalize]{cleveref}
\crefname{section}{Sec.}{Secs.}
\Crefname{section}{Section}{Sections}
\Crefname{table}{Table}{Tables}
\crefname{table}{Tab.}{Tabs.}

\def\cvprPaperID{8177} 
\def\confName{CVPR}
\def\confYear{2022}

\begin{document}

\title{Unsupervised Representation Learning for Binary Networks\\ by Joint Classifier Learning}

\makeatletter
\def\thanks#1{\protected@xdef\@thanks{\@thanks
        \protect\footnotetext{#1}}}
\makeatother

\author{
\vspace{0.2em} 
\hspace{2em}Dahyun Kim\textsuperscript{1,2}\hspace{6em}
\hspace{2em}Jonghyun Choi\textsuperscript{2,3,$\dagger$}
\thanks{\hspace{-1.5em}This work was done while DK and JC were an intern and an AI technical advisor at NAVER AI Lab., respectively. $^\dagger$ indicates corresponding author.}\\
{\textsuperscript{1}Upstage AI Research\hspace{2em}
\textsuperscript{2}NAVER AI Lab.\hspace{2em}
\textsuperscript{3}Yonsei University}\\
{\tt\small {kdahyun@upstage.ai \hspace{8em} jc@yonsei.ac.kr}}
}

\maketitle

\begin{abstract}
Self-supervised learning is a promising unsupervised learning framework that has achieved success with large floating point networks.
But such networks are not readily deployable to edge devices.
To accelerate deployment of models with the benefit of unsupervised representation learning to such resource limited devices for various downstream tasks, we propose a self-supervised learning method for binary networks that uses a moving target network.
In particular, we propose to jointly train a randomly initialized classifier, attached to a pretrained floating point feature extractor, with a binary network.
Additionally, we propose a feature similarity loss, a dynamic loss balancing and modified multi-stage training to further improve the accuracy, and call our method \emph{\method}.
Our empirical validations over five downstream tasks using seven datasets show that \method outperforms self-supervised baselines for binary networks and sometimes outperforms supervised pretraining. Code is availabe at {\small\url{https://github.com/naver-ai/burn}}.
\end{abstract}

\section{Introduction}
\label{sec:intro}
Self-supervised learning (SSL) has achieved great success with floating point (FP) networks in recent years~\cite{chen2021exploring, goyal2021self, tian2021divide, zbontar2021barlow,Li_2021_CVPR,Cai_2021_CVPR,Ericsson_2021_CVPR,tian2021understanding,ermolov2021whitening,tian2020makes,chen2020big,grill2020bootstrap,caron2020unsupervised, he2020momentum}. 
Models learned by SSL methods perform \emph{on par} with or even outperform the ones learned by supervised pretraining by the help of large scale unlabeled data in a number of downstream tasks such as image classification~\cite{compress, caron2020unsupervised}, semi-supervised fine-tuning~\cite{grill2020bootstrap, caron2020unsupervised, chen2020big} and object detection~\cite{he2020momentum}.
While recent works~\cite{chen2020big, grill2020bootstrap, caron2020unsupervised,he2020momentum} from resourceful research groups have shown that the gains from SSL scale up with model size and/or dataset size used for pretraining, there is little work where the resulting pretrained models are small in size, \ie, quantized. 
SSL for such small models is important since it could expedite the AI deployment for a wide range of applications onto models with high efficiency in computational and memory costs, and energy consumption~\cite{binsolar}.
%
At the extreme of resource constrained scenarios, binary networks exhibit superior efficiency and the accuracy is being significantly improved~\cite{Rastegari2016XNORNetIC, lin2017towards,liu2018bi, liu2020reactnet, Martinez2020Training, Bulat2020BATSBA, kimSC2020BNAS, bulat2021highcapacity}.
Thus, developing an SSL method for binary networks could further accelerate the deployment of models to edge devices for various downstream tasks, yet is seldom explored.

\begin{figure}[t!]
    \centering
    \resizebox{0.95\linewidth}{!}{
    \includegraphics{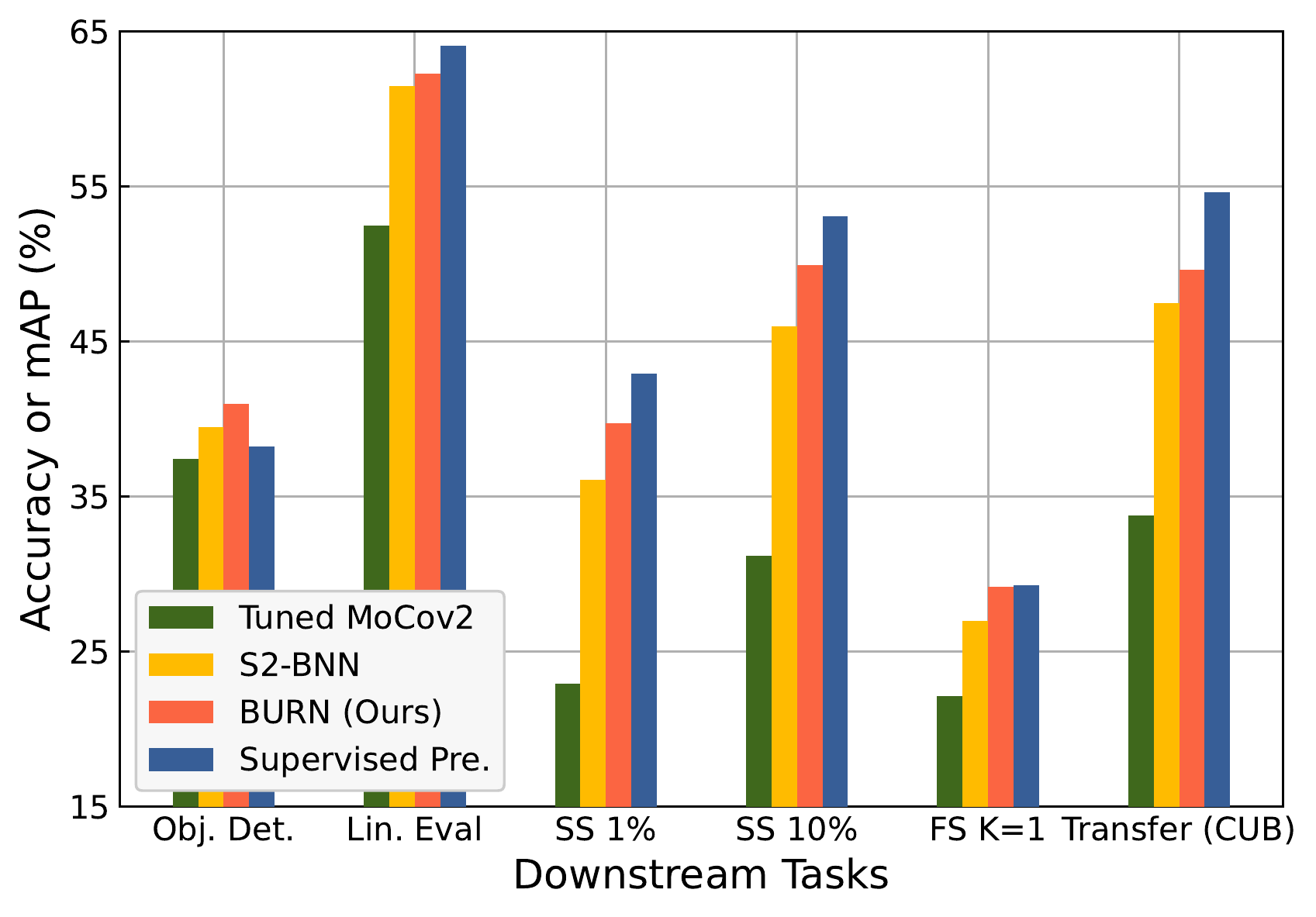}
    }
    \vspace{-1.2em}
    \caption{Comparison of various representation learning methods on multiple downstream tasks (pretrained with ImageNet). `Obj. Det.' refers to object detection, `Lin. Eval' refers to linear evaluation, 'SS 1/10\% refers to semi-supervised fine-tuning with 1 or 10\% data respectively,`FS K=1' refers to few-shot learning with 1 shot, and `Transfer (CUB)' means transfer learning to the CUB dataset. `Tuned MoCov2' and `S2-BNN' are SSL methods from \cite{shen2021s2}. Proposed \method outperforms all comparable methods in various tasks, and even the Supervised Pre. in certain tasks. } 
    \vspace{-1em}
    \label{fig:teaser}
\end{figure}

Providing additional supervisory signals with a pretrained FP network by using the KL divergence loss between the softmax outputs from the classifiers of the FP target network and binary network, which we denote as \emph{`supervised KL div.'}, has become a popular and effective method for training binary networks~\cite{Martinez2020Training, liu2020reactnet, bulat2021highcapacity, Bulat2020BATSBA}.
Recently, \cite{shen2021s2} propose an unsupervised representation learning method for binary networks based on the supervised KL div. method.
To extract meaningful softmax probabilities from the FP network, they pretrain the classifier as well as the feature extractor using SSL.
Then, the FP network is completely frozen when used as the target network, which could lead to stale targets~\cite{grill2020bootstrap} or be dependent on the pretraining dataset used for the fixed FP network being similar to the dataset used for training the binary network.

Thus, to avoid the potential pitfalls of a fixed target, we are motivated to develop an SSL method for binary networks that uses a moving FP network as the target, similar to other SSL methods~\cite{grill2020bootstrap,he2020momentum,chen2021exploring, chen2020improved}, and call our method {\bf B}inary {\bf U}nsupervised {\bf R}epresentatio{\bf N} learning or {\bf \method}.
Specifically, we first construct the FP target network by combining a fixed FP feature extractor pretrained in an SSL manner and a randomly initialized FP classifier.
We then use the outputs of the randomly initialized FP classifier as targets for the binary network and \emph{jointly optimize both the FP classifier and the binary network}, using the KL divergence loss, to keep updating the FP network overtime.
But the gradients provided by the randomly initialized FP classifier could have unexpectedly large magnitudes, especially during early training phase.
To alleviate this problem, we additionally propose to enforce feature similarity across both precision, providing stable gradients that bypass the randomly initialized classifier.
As relative importance of the feature similarity loss decreases as the FP classifier gets jointly trained to provide less random targets, we further propose to \emph{dynamically balance} the KL divergence term and the feature similarity term in the loss function.
Finally, we modify the multi-stage training scheme~\cite{Martinez2020Training} for \method to further improve performance.

We conduct extensive empirical validations with a wide variety of downstream tasks such as object detection on Pascal VOC, linear evaluation on ImageNet, semi-supervised fine-tuning on ImageNet with 1\% and 10\% labeled data, SVM classification and few-shot SVM classification on Pascal VOC07, and transfer learning to various datasets such as CIFAR10, CIFAR100, CUB-200-2011, Birdsnap, and Places205. 
In the validations, the binary networks trained by our method outperforms other SSL methods by large margins (see Fig.~\ref{fig:teaser} and Sec.~\ref{sec:down_task}).

We summarize our contributions as follows:
\vspace{-0.5em}
\begin{itemize}[leftmargin=10pt]
\setlength\itemsep{-0.1em}
   \item We propose a novel SSL method for binary networks that uses a jointly trained FP classifier to obtain targets that can adapt overtime to the current training scenario.
   \item We propose to use a feature similarity loss and dynamic balancing with modified multi-stage training to significantly improve the accuracy.
   \item Our \method outperforms prior arts by large margins on a wide variety of downstream tasks.
   \item We analyze our proposed \method by in-depth investigations.
\end{itemize}

\section{Related Work}
\label{sec:related}

\subsection{Self-Supervised Representation Learning}
To reduce the annotation cost for representation learning, self-supervised representation learning (SSL) methods including ~\cite{chen2021exploring, goyal2021self,tian2021divide,zbontar2021barlow,tian2020makes,tian2019contrastive, chen2020simple,he2020momentum,chen2020big} and many more have been shown to be effective, with the Info-NCE loss~\cite{oord2018representation} being a popular choice for many works.
These methods use the instance discrimination task as the pretext task which aims to pull instances of the same image closer and push instances of different images farther apart~\cite{wu2018unsupervised, oord2018representation}.
Different to these methods, \cite{grill2020bootstrap,caron2020unsupervised, li2021prototypical, wei2021co, fang2021seed, compress} use feature regression with an EMA target~\cite{grill2020bootstrap}, matching cluster assignments~\cite{caron2020unsupervised, li2021prototypical}, or matching similarity score distributions~\cite{wei2021co, fang2021seed, compress} as the pretext task.
We compare with BYOL~\cite{grill2020bootstrap} and SWAV~\cite{caron2020unsupervised} as they show high performance and have similarities with other SSL methods~\cite{chen2021exploring, tian2020makes, chen2020big}.
However, while these methods show promising results for large FP models, they do not consider resource constrained scenarios which are more practical, \textit{e.g.}, quantized models with smaller complexity.


\subsection{Binary Networks}
\label{sec:related_bin}
At the extreme of quantized models, numerous works on binary networks~\cite{Rastegari2016XNORNetIC,lin2017towards,liu2018bi, liu2020reactnet, Martinez2020Training, Bulat2020BATSBA, kimSC2020BNAS, bulat2021highcapacity, rotatedbinary, Qin_2020_CVPR, Han2020TrainingBN, meng2020training, Kim2020BinaryDuo} have been proposed.
These include searching architectures for binary networks~\cite{kimSC2020BNAS,kimbnasv2, Bulat2020BATSBA}, using specialized activation functions~\cite{liu2020reactnet}, and object detection using the information bottleneck principle~\cite{wang2020bidet}.
Note that previous works mostly focused on the supervised training set-up.

Among many proposals, two stand out as the state-of-the-art binary network backbones due to their strong empirical performance: ReActNet~\cite{liu2020reactnet} and High-Capacity Expert Binary Networks (HCEBN)~\cite{bulat2021highcapacity}.
\cite{liu2020reactnet} propose to learn thresholds for binarization by the RSign and RPReLU activation functions.
\cite{bulat2021highcapacity} use multiple experts for conditional computing and increase the representation capacity of binary networks without increasing the operation count.

Recently, both `supervised KL div.' method and the multi-stage training scheme~\cite{liu2020reactnet, Martinez2020Training} have become popular for training binary networks.
The supervised KL div. method uses a pretrained FP network to provide targets for the KL div. loss in training binary networks.
The multi-stage training scheme trains a binary network in multiple stages, where more parts of the network are binarized.
%
\begin{figure*}[t!]
    \centering
    \resizebox{0.95\linewidth}{!}{
    \includegraphics{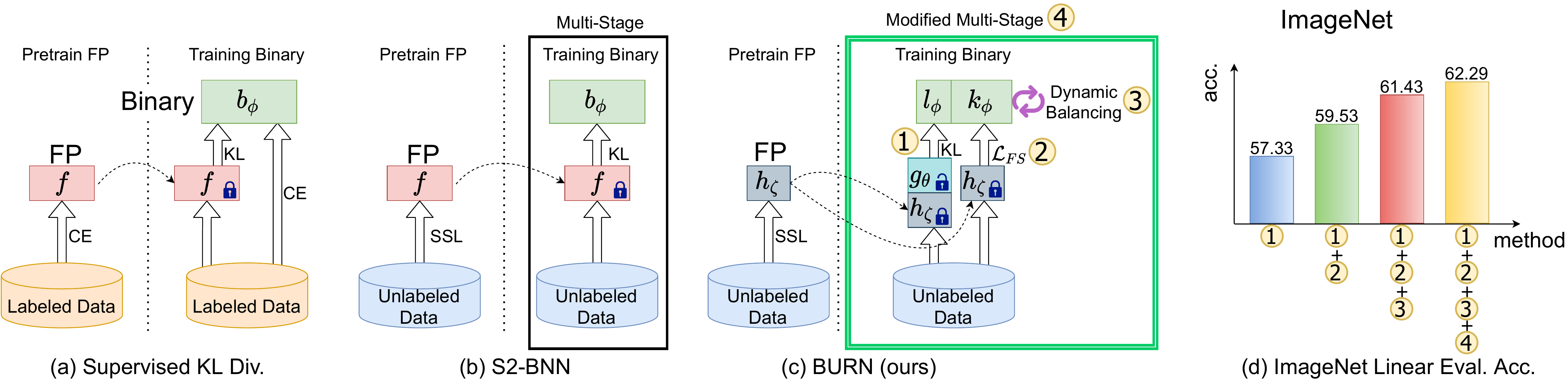})
    }
    \vspace{-1em}
    \caption{Illustrative comparison of Supervised KL div. method~\cite{Martinez2020Training, liu2020reactnet}, S2-BNN~\cite{shen2021s2} and proposed \method, and linear evaluation accuracy of ablated models on ImageNet. $f$ is the FP network, $h_\zeta$ is the FP feature extractor, $g_\theta$ is the trainable classifier, and $k_\phi, l_\phi$ are the binary feature extractor and classifier decoupled from the binary network $b_\phi$. $f$ and $h_\zeta$ are initialized with pretrained weights as indicated by the dotted arrow. The `locked' icons in $f$ and $h_\zeta$ indicate they are not trained while the `unlocked' icon in $g_\theta$ indicates that it is trained. Our baseline (\numcircledmod{1}) already achieves a relatively high top-1 accuracy of $57.33\%$ on ImageNet and our proposed components provide noticeable gains. For details, please see Sec.~\ref{sec:naive_baseline} for \numcircledmod{1}, Sec.~\ref{sec:augmented_loss} for \numcircledmod{2}, Sec.~\ref{sec:balancing} for \numcircledmod{3}, and Sec.~\ref{sec:multi} for \numcircledmod{4}.} 
    \vspace{-1em}
    \label{fig:setting}
\end{figure*}
The most related work is the recently published S2-BNN\cite{shen2021s2} which utilize the supervised KL div. method to train binary networks in an SSL manner.
They use unlabeled data when training the binary network with only the KL divergence loss between the binary and the pretrained and frozen FP networks.
S2-BNN shows good performance but the frozen FP network as a target may be limiting \ie, many works in the SSL literature suggest that a \emph{changing} target is effective~\cite{grill2020bootstrap,chen2020improved,he2020momentum, chen2021exploring}.
We extensively compare with S2-BNN.

In contrast, we aim to develop an unsupervised representation learning method for binary networks which uses a FP network that is changing as the target. 

\section{Approach}
\label{sec:approach}

The supervised KL div. method is an effective method to train binary networks~\cite{Martinez2020Training, liu2020reactnet} that utilizes a FP network pretrained with labeled data.
But, as we are interested in the self-supervised learning with no access to labeled data at any time during training, the supervised KL div. is not applicable.
Recently, S2-BNN~\cite{shen2021s2} propose to use the supervised KL div. for unsupervised learning of binary networks.
They pretrain the classifier and the feature extractor of the FP network to obtain meaningful softmax probabilities and use a completely fixed FP network as the target.
In contrast, we propose an unsupervised representation learning method for binary networks that uses a changing FP network as the target such that the FP network can adapt to the current dataset and binary network to provide more useful targets overtime.
We illustrate the supervised KL div. method~\cite{Martinez2020Training,liu2020reactnet}, S2-BNN~\cite{shen2021s2}, and our proposal in Fig.~\ref{fig:setting}.

Specifically, instead of using softmax outputs from a fixed pretrained FP network~\cite{shen2021s2}, we propose to use softmax outputs from \emph{a randomly initialized classifier} attached to a pretrained FP feature extractor, and jointly train the classifier with the binary network using the KL divergence loss.    
As the supervision from the untrained classifier makes gradients with unexpectedly high magnitudes, we subdue gradients by proposing an additional feature similarity loss across precision.
We propose to use a dynamic balancing scheme between the loss terms to better balance the KL divergence and feature similarity losses and employ a modified multi-stage training~\cite{Martinez2020Training} to improve learning efficacy.

\subsection{Joint Classifier Training as Moving Targets}
\label{sec:naive_baseline}

Grill \etal~\cite{grill2020bootstrap} show that even when a randomly initialized exponential moving average (EMA) network is used as the target network, the online network improves by training with it.
One possible reason for the improvement is that the randomly initialized target network is also updated in an EMA manner during training, improving it gradually.
Motivated by this, we conjecture whether a randomly initialized classifier combined with a pretrained FP feature extractor can be used as a \emph{moving target network} for training binary networks.
To gradually improve the target network, we jointly train the classifier of the target network and the binary network.
Note that training just the classifier can improve the target network as is shown in the SSL literature~\cite{chen2021exploring, goyal2021self,tian2021divide,zbontar2021barlow,tian2019contrastive, chen2020simple,he2020momentum,tian2020makes,chen2020big}.
We discuss other moving targets, \eg, the EMA target~\cite{grill2020bootstrap} or the momentum encoder~\cite{he2020momentum} for binary networks in Sec.~\ref{sec:mov_targ_choice}.

The joint training of the randomly initialized classifier is depicted in \numcircledmod{1} in Fig.~\ref{fig:setting}-(b).
Specifically, instead of a fixed FP network $f(\cdot)$, the randomly initialized and trainable classifier $g_{\theta}(\cdot)$ and the pretrained and fixed FP feature extractor $h_{\zeta}(\cdot)$ are combined to create the target network.
Then, we use the outputs of $g_{\theta}(\cdot)$ as targets for training the binary network $b_{\phi}(\cdot)$.
Our objective is to minimize the KL divergence between the outputs of $g_{\theta}(\cdot)$ and $b_{\phi}(\cdot)$ as:
\begin{equation}
    \min_{\theta, \phi} \mathbb{E}_{x \sim \mathcal{D}} [ \mathcal{L}_{KL}(g_{\theta}(h_{\zeta}(x)), b_{\phi}(x))],
\label{eq:baseline}
\end{equation}
where $x$ is a sample from the dataset $\mathcal{D}$ and $\mathcal{L}_{KL} = D_{KL}(\cdot,\cdot)$ is the KL divergence between the outputs of $g_{\theta}(\cdot)$ and $b_{\phi}(\cdot)$.
%
However, the softmax outputs from the classifier would be close to random early on.
Thus, using the random outputs as the only target for the binary network, especially in early training, could result in noisy gradients.

\subsection{Stabilize Gradients by Feature Similarity Across Precision} 
\label{sec:augmented_loss}
\begin{figure}[t!]
    \centering
    \resizebox{0.96\linewidth}{!}{
    \includegraphics{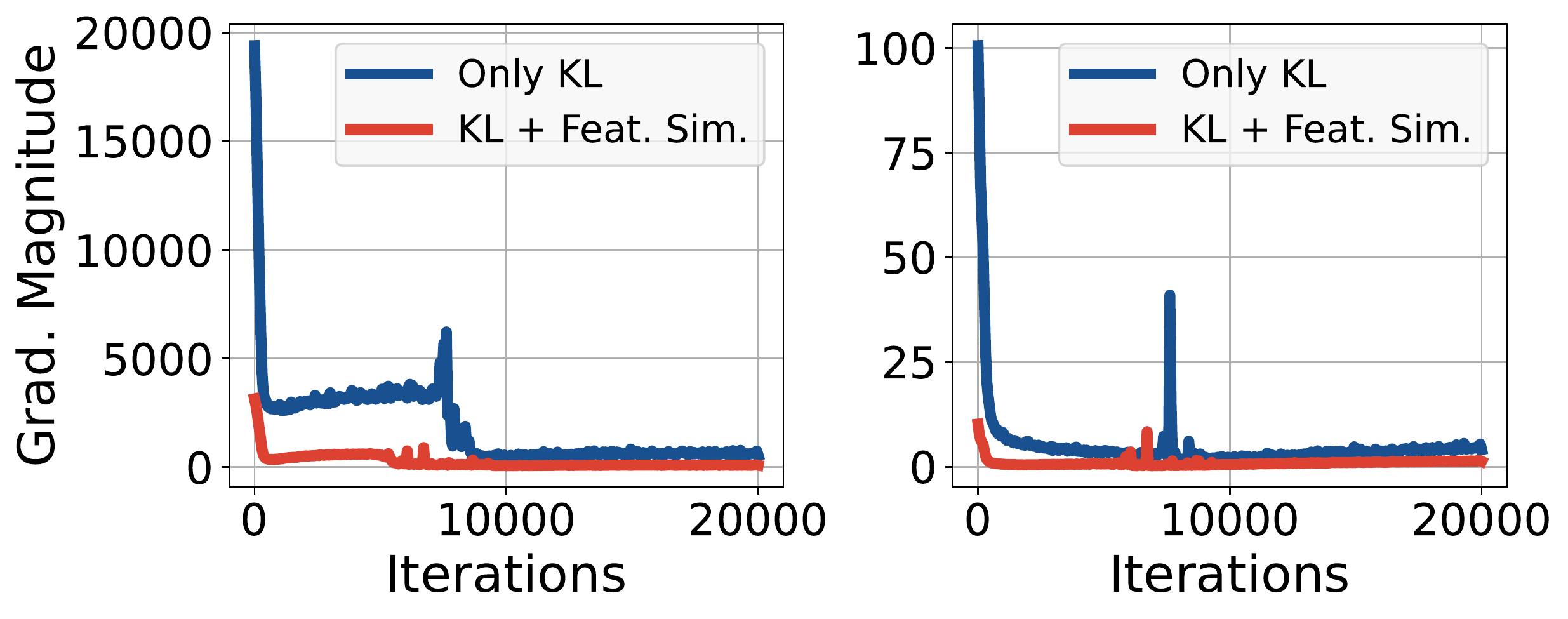}
    }\\
    {\footnotesize \hspace{5em}(a) Classifier \hspace{7em}(b) Feature extractor}
    \vspace{-0.5em}
    \caption{Gradient magnitude for the binary classifier (a) and the binary feature extractor (b) during early training with and without $\mathcal{L}_{FS}$ for pretraining on ImageNet. With only KL, the gradients of the classifier is extremely large and this carries over to the feature extractor. 
    Additionally, we observe intermediate spikes for both the classifier and the feature extractor. The addition of $\mathcal{L}_{FS}$ significantly lowers the gradient magnitudes of the classifier as well as the feature extractor at early iterations. Additionally, the surges in gradient magnitudes are also subdued.}
    \vspace{-0.5em}
    \label{fig:grad_spike}
\end{figure}
To alleviate the issue of unreliable gradients from the randomly initialized classifier being the only target, particularly in the early the training, we propose an additional loss term that enforces feature similarity between the target and the binary network.
Specifically, $g_{\theta}(\cdot)$ is largely updated (due to the fixed feature extractor) in the early phase of the joint training. 
As the binary classifier uses the quickly changing $g_{\theta}(\cdot)$ as a target to transfer knowledge from, the binary classifier might receive large gradients.
To address the potentially undesirably large gradients caused by using the randomly initialized classifier as the only target, we propose to augment an additional loss term that bypasses the classifier. 
We call it the \emph{feature similarity loss}.

Specifically, we use the cosine distance between the feature vectors from the FP and binary feature extractors as the feature similarity loss; $\mathcal{L}_{FS}(v_1, v_2) = 1-{\langle v_1, v_2 \rangle \over \|v_1\|_2 \cdot \|v_2\|_2}$ for smoothness and a bounded nature to prevent large gradients.
The cosine distance (or $1-$the cosine similarity) is widely used in representation learning~\cite{grill2020bootstrap, xiao2021what, he2020momentum, chen2020simple} (discussion on other choices for $\mathcal{L}_{FS}$ is in Sec.~\ref{sec:exp_choice_fs}).

Augmenting the cosine distance to the KL divergence loss, we can write our new optimization problem as:
\begin{equation}
\begin{split}
    \min_{\theta, \phi} \mathbb{E}_{x \sim \mathcal{D}} [ (1-\lambda) &\mathcal{L}_{KL}(g_{\theta}(h_{\zeta}(x)), l_{\phi}(k_{\phi}(x)))\\
    +&\lambda \mathcal{L}_{FS}(h_{\zeta}(x), k_{\phi}(x))],
\end{split}
\label{eq:aug}
\end{equation}
where the binary network $b_{\phi}(\cdot)$ is decoupled into $k_{\phi}(\cdot)$ the binary feature extractor and the classifier $l_{\phi}(\cdot)$, $\lambda$ is a static balancing factor, and $\mathcal{L}_{FS}(\cdot, \cdot)$ is the feature similarity loss.

The new loss provides additional supervisory signals from the feature extractor of the FP network.
Since the feature extractor of the FP network is pretrained and fixed, it provides stationary and stable targets as opposed to the randomly initialized classifier.
Empirically, we observe the gradient of the binary classifier and feature extractor with and without $\mathcal{L}_{FS}$ in Fig~\ref{fig:grad_spike}.
Note that with only KL, the gradients of the binary classifier are extremely large; it starts at roughly $20,000$ then drops to roughly $3,000$ for some iterations and finally drops to a small value around roughly $9,000$ iterations.
In addition, there is a surge in gradient magnitude around $7,500$ iterations.
The binary feature extractor also shows a similar trend where the gradients exhibit a sudden spike at around $7,500$ iterations.
Both very high magnitudes of the gradients at the start and the sudden spikes, occurring after some iterations, would harm training stability~\cite{zhang2019gradient, chen2020understanding}.
However, as shown in the figures, addition of the proposed $\mathcal{L}_{FS}(\cdot, \cdot)$ significantly reduces the gradient magnitudes of the binary classifier and the feature extractor at early iterations as well as the surges throughout the training, which leads to better training efficacy and accuracy.

\subsection{Dynamic Balancing of $\lambda$}
\label{sec:balancing}
As $g_{\theta}$ is gradually updated, it provides more meaningful targets and $\mathcal{L}_{FS}$ becomes less important.
Thus, we propose a temporally \emph{dynamic balancing} strategy to replace the static balancing factor $\lambda$ in Eq.~\ref{eq:aug} by a smooth cosine annealing similar to how \cite{grill2020bootstrap} annealed the momentum value as:
%
\begin{equation}
    \lambda(t) = \lambda_{T_{max}} - (\lambda_{T_{max}} - \lambda_{0}) \cdot (\cos(\pi t / T_{max}) + 1) / 2,
\label{eq:dynamic_lambda}
\end{equation}
where $\lambda_{0}$ and $\lambda_{T_{max}}$ are the initial and final values of $\lambda(t)$, $T_{max}$ is the maximum training iteration and $t$ is the current training iteration.
Thus, $\lambda(t)$ will start at $\lambda_{0}$ then gradually decay to $\lambda_{T_{max}}$, emphasizing the cosine distance more at the beginning and less as the learning progresses.
Discussion on other choices of $\lambda(t)$ are in Sec.~\ref{sec:choice_of_lamdbda}.
%

Finally, our optimization problem can be rewritten as:
\begin{equation}
\begin{split}
    \min_{\theta, \phi} \mathbb{E}_{x \sim \mathcal{D}}[(1-\lambda(t))&\mathcal{L}_{KL}(g_{\theta}(h_{\zeta}(x)), l_{\phi}(k_{\phi}(x)))\\ +\lambda(t)&\mathcal{L}_{FS}(h_{\zeta}(x), k_{\phi}(x))].
\end{split}
\label{eq:final_obj}
\end{equation}

\begin{algorithm}[t]
\caption{{\bf B}inary {\bf U}nsupervised {\bf R}epresentatio{\bf N} learning (\method)}
\label{algo:ripl}
\begin{algorithmic}[1]
    \Function{BURN}{$\mathcal{D}$, $t$, $\zeta$, $h_{\zeta}$, $g_{\theta}$, $k_\phi$, $l_{\phi}$}
    \State $\theta, \phi \leftarrow \text{Pretrain($\mathcal{D}$, $t$, $\zeta$, $h_{\zeta}$, $g_{\theta}$, $k_\phi$, $l_{\phi}$, STAGE1)}$
    \State $W \leftarrow \{\zeta\} \bigcup \{\theta, \phi\}$
    \State $\theta, \phi \leftarrow \text{Pretrain($\mathcal{D}$, $t$, $W$, $h_{\zeta}$, $g_{\theta}$, $k_\phi$, $l_{\phi}$, STAGE2)}$
    \State \textbf{return} $k_\phi$
    \EndFunction
    
    \Function{Pretrain}{$\mathcal{D}$, $t$, $W$, $h_{\zeta}$, $g_{\theta}$, $k_\phi$, $l_{\phi}$, F}
    \If{F is STAGE1}
    \State $k_\phi, l_\phi \leftarrow \text{Binarize Activations}$
     \State $h_{\zeta} \leftarrow W$ {\small\color{azure}\Comment{Load pretrained weights}}
    \Else
        \State $k_\phi, l_\phi \leftarrow \text{Binarize Activations and Weights}$
        \State $h_{\zeta}, g_{\theta}, k_{\phi}, l_{\phi} \leftarrow W$ {\small\color{azure}\Comment{Load pretrained weights}}
    \EndIf
    \State $x = \text{RandomSelect}(\mathcal{D})$ {\small\color{azure}\Comment{Sample $x$ $\sim$ $\mathcal{D}$}}
    \State $v_1, v_2 = h_{\zeta}(x), k_{\phi}(x)$ {\small \color{azure}\Comment{Feature vectors $v_1, v_2$}}
    \State $p_1, p_2 = g_{\theta}(v_1), l_{\phi}(v_2)$ {\small \color{azure}\Comment{Softmax Probabilities $p_1, p_2$}}
    \State $\mathcal{L}_{\zeta,\theta, \phi} = \text{AugmentedLoss}(v_1, v_2, p_1, p_2, t)$
    \State $\theta \leftarrow \text{Optimizer}(\nabla_\theta \mathcal{L}_{\zeta,\theta, \phi}, \eta)$ {\small \color{azure}\Comment{Update $\theta$}}
    \State $\phi \leftarrow \text{Optimizer}(\nabla_\phi \mathcal{L}_{\zeta,\theta, \phi}, \eta)$ {\small \color{azure}\Comment{Update $\phi$}}
    \State \textbf{return} $\theta, \phi$
    \EndFunction

\Function{AugmentedLoss}{$v_1$, $v_2$, $p_1$, $p_2$, $t$}
    \State $\mathcal{L}_{KL} = \mathcal{D}_{KL}(p_2 \|\| p_1)$ {\small \color{azure}\Comment{KL Divergence}}
    \State $\mathcal{L}_{FS} = 1-{\langle v_1, v_2 \rangle \over \|v_1\|_2 \cdot \|v_2\|_2}$ {\small \color{azure}\Comment{Cosine Distance}}
    \State $\lambda(t) = \lambda_{T} - (\lambda_{T} - \lambda_{0}) \cdot (\cos(\pi t / T) + 1) / 2$ {\small \color{azure}\Comment{Eq.~\ref{eq:dynamic_lambda}}}
    \State $\mathcal{L} = (1-\lambda(t))\cdot\mathcal{L}_{KL} + \lambda(t)\cdot\mathcal{L}_{aug}$ {\small \color{azure}\Comment{Eq.~\ref{eq:final_obj}}}
    \State \textbf{return} $\mathcal{L}$
    \EndFunction
\end{algorithmic}
\end{algorithm}

\subsection{Modified Multi-Stage Training for \method}
\label{sec:multi}
The multi-stage training ~\cite{Martinez2020Training,Bulat2020BATSBA,liu2020reactnet} is known to be effective in training binary networks.
It trains the network with only binarized activations in the first stage.
Then, it uses the trained weights of the partially binarized network as initial values for training the fully binarized network, \ie, binarized weights and activations, in the second stage.
Unfortunately, we cannot use this strategy as the binary networks converge quickly thanks to the good initial values learned in the first stage~\cite{Martinez2020Training} whereas the randomly initialized FP classifier $g_{\theta}$ does not converge as quickly as the binary network.
This discrepancy in the convergence speeds of the binary network and the FP classifier harms training efficacy.

To apply the multi-stage training to \method, we modify it to give good initial points to the FP classifier as well as the binary network.
Specifically, we initialize $g_{\theta}$ in the second stage with the weights of $g_{\theta}$ obtained in the first stage, similar to the binary network.
As a result, $g_{\theta}$ starts from a good initial point and converges quickly to provide useful targets.

We describe the full algorithm of {\bf \method} in Alg.~\ref{algo:ripl}

\section{Experiments}
\label{sec:exp}


\paragraph{Experimental Details.}
Following \cite{xiao2021what, lee2021imix, debiasedcont, hardnegmix, zhao2021what, wang2020hypersphere}, we use ImageNet~\cite{krizhevsky2012imagenet} for pretraining.
We use 1) object detection, 2) linear evaluation, 3) semi-supervised fine-tuning, 4) full-shot and few-shot image classification using SVM, and 5) transfer learning via linear evaluation for downstream tasks.
We strictly follow the SSL evaluation protocols of the downstream tasks~\cite{goyal2019scaling, he2020momentum,chen2020big, chen2020simple}.
Downstream task and implementation details are in the supplement.
Experiments were partly based on NAVER Smart Machine Learning (NSML) platform~\cite{kim2018nsml,sung2017nsml}.
Code is availabe at \url{https://github.com/naver-ai/burn}.

\vspace{-1em}\paragraph{Baselines.}
We pretrain the ReActNet-A backbone with BYOL~\cite{grill2020bootstrap}, SWAV~\cite{caron2020unsupervised}, tuned MoCov2~\cite{shen2021s2}, and S2-BNN~\cite{shen2021s2} as our SSL baselines.
We also show supervised pretraining, \ie, `Supervised Pre.'.
More comparisons to SimCLRv2~\cite{chen2020big} and InfoMin~\cite{tian2020makes} are in the supplement.
We pretrain the models for $200$ epochs in all methods.

\begin{table}
\centering
    \centering
    \resizebox{0.9\linewidth}{!}{
    \begin{tabular}{cccc}
    \toprule
        Method          & mAP (\%) & AP50 (\%) & AP75 (\%)  \\ \midrule
        Supervised Pre.&     38.22 & 68.53 & 37.65    \\  \midrule
        SWAV~\cite{caron2020unsupervised}            &    37.22 & 67.47 & 35.91          \\ 
        BYOL~\cite{grill2020bootstrap}           &     36.92 & 67.13 & 35.65       \\
        Tuned MoCov2~\cite{shen2021s2}    & 37.42 & 67.30 & 36.37 \\
        S2-BNN~\cite{shen2021s2} & 39.50 & 70.09 & 39.15 \\
        \cellcolor{Orange!20}\method (Ours)&\cellcolor{Orange!20}{\bf41.00} & \cellcolor{Orange!20}{\bf 70.91} & \cellcolor{Orange!20}{\bf 41.45}\\\bottomrule
    \\
    \end{tabular}
    }
    \vspace{-1.5em}
    \caption{Object detection (mAP, AP50 and AP75) on Pascal VOC after pretraining. \method outperforms all the compared methods including supervised pretraining (`Supervised Pre.').} 
    \vspace{-1em}
    \label{tab:detection}
\end{table}

\begin{table*}[t!]
    \centering
    \resizebox{0.7\linewidth}{!}{
\begin{tabular}{cccccc}
\toprule
 \multirow{3.5}{*}{Method}  & \multirow{2}{*}{Linear Eval.} & \multicolumn{4}{c}{Semi-Supervised Fine-tuning}\\
 && \multicolumn{2}{c}{1\% Labels} & \multicolumn{2}{c}{10\% Labels} \\ 
 \cmidrule(lr){2-2} \cmidrule(lr){3-4}\cmidrule(lr){5-6}
 & Top-1 (\%) & Top-1 (\%) & Top-5 (\%) & Top-1 (\%) & Top-5 (\%) \\ 
 \cmidrule(lr){1-1}\cmidrule(lr){2-2}\cmidrule(lr){3-4}\cmidrule(lr){5-6}
    Supervised Pre. &64.10   &42.96&	69.10&				53.07&	77.40 \\
 \cmidrule(lr){1-1}\cmidrule(lr){2-2} \cmidrule(lr){3-4}\cmidrule(lr){5-6}
    SWAV& 49.41&  24.66&	46.57	&			33.83&	57.81      \\
    BYOL &  49.25&23.05&	43.90	&			34.66&	58.78    \\ 
    Tuned MoCov2 & 52.50& 22.96 &  45.12& 31.18 &55.64 \\
    S2-BNN & 61.50 & 36.08 & 61.83 & 45.98 & 71.11 \\
    \cellcolor{Orange!20}\method (Ours)  & \cellcolor{Orange!20}{\bf62.29}&\cellcolor{Orange!20}{\bf39.75}&\cellcolor{Orange!20}{\bf67.13}	&\cellcolor{Orange!20}{\bf49.96}&\cellcolor{Orange!20}{\bf75.52}       \\
\bottomrule
\end{tabular}
}
\vspace{-0.5em}
\caption{Linear evaluation (top-1) and semi-supervised fine-tuning (1\% labels or 10\% labels) on ImageNet after pretraining. \method outperforms all other SSL methods by large margins across for both the linear evaluation and semi-supervised  fine-tuning.}
\vspace{-0.5em}
\label{tab:cls_semi}
\end{table*}

\begin{table*}[t!]
    \centering
    \resizebox{0.95\linewidth}{!}{
        \begin{tabular}{cccccccccc}
        \toprule
        Method          & k = 1& k = 2 & k = 4 & k = 8 & k = 16 & k = 32 & k = 64 & k = 96 & Full\\ \midrule
        Supervised Pre.     &29.28$\pm$ 0.94&	36.46$\pm$ 2.97&	49.67$\pm$ 1.20&	56.99$\pm$ 0.67&	64.68$\pm$0.89&	70.08$\pm$ 0.58&	73.49$\pm$ 0.53&	74.96$\pm$ 0.17 & 77.47\\ \midrule
        SWAV &  22.97 $\pm$ 1.21&	27.91$\pm$ 2.37&	37.91$\pm$1.11&	44.5$\pm$ 1.51&	52.79$\pm$ 0.81&	59.15$\pm$ 0.62&	64.38$\pm$ 0.59&	66.72$\pm$ 0.19  & 71.23 \\
        BYOL  & 23.45 $\pm$ 0.76&	28.04$\pm$ 2.40&	38.09$\pm$ 1.07&	44.69$\pm$ 1.66&	51.5$\pm$ 0.90&	57.44$\pm$0.24&	62.07$\pm$ 0.28&	64.37$\pm$ 0.13 & 69.16\\
        Tuned MoCov2 & 22.12 $\pm$ 0.74 & 27.45 $\pm$ 2.06& 36.81 $\pm$ 0.82& 43.19 $\pm$1.4& 51.93 $\pm$0.84& 57.95 $\pm$ 0.62&63.07 $\pm$ 0.43&65.15 $\pm$0.05& 69.73\\
        S2-BNN & 27.00$\pm$1.54 & 33.39 $\pm$ 2.72& 46.31 $\pm$ 2.11 & 54.14 $\pm$1.32&61.86 $\pm$ 1.14&68.01 $\pm$ 0.41&71.89 $\pm$ 0.44&73.55$\pm$ 0.29 & 76.49 \\ 
        \cellcolor{Orange!20} \method (Ours) &\cellcolor{Orange!20}{\bf29.20$\pm$1.51}&\cellcolor{Orange!20}{\bf36.14 $\pm$2.15}&\cellcolor{Orange!20}{\bf48.49 $\pm$ 1.08}&\cellcolor{Orange!20}{\bf55.12 $\pm$ 1.59}&\cellcolor{Orange!20}{\bf62.36 $\pm$ 1.01}&\cellcolor{Orange!20}{\bf68.10 $\pm$ 0.3}&\cellcolor{Orange!20}{\bf72.1 $\pm$ 0.39}&\cellcolor{Orange!20}{\bf74.06 $\pm$ 0.18}&\cellcolor{Orange!20}{\bf77.49} \\ \bottomrule
        \\
        \end{tabular}
        }
        \vspace{-1.5em}
    \caption{SVM classification (mAP) for the few-shot and full-shot settings on VOC07 after pretraining. 
    \method outperforms all other SSL methods by large margins and performs on par with supervised pretraining on both settings. The number of shots ($k$) is varied from 1 to 96. We report the averaged performance over 5 runs with the standard deviation.}
    \vspace{-1em}
\label{tab:fewshot}
\end{table*}

\subsection{Results on Downstream Tasks}
\label{sec:down_task}

We evaluate our method along with prior arts in various downstream tasks.
We denote the best results except for supervised pretraining in each table in {\bf bold.}

\vspace{-1em}\paragraph{Object Detection.}
We first conduct object detection (mAP (\%), AP50 (\%) and AP75 (\%)) on Pascal VOC and summarize the results in Table~\ref{tab:detection}.
Once the feature extractor is pretrained, we use the pretrained weights as initial weights for fine-tuning a detection pipeline.
\method outperforms all other methods, including supervised pretraining, in all three metrics.
We believe one of the reasons for the performance of \method is that it utilizes a FP network trained in an SSL manner that mostly learned low- and mid-level features~\cite{zhao2021what} which would help object detection. 

\vspace{-1em}\paragraph{Linear Evaluation.}
We then conduct linear evaluation (top-1) on ImageNet and summarize the results in Table~\ref{tab:cls_semi}.
Once the binary feature extractor is pretrained, it is frozen and only the attached classifier is trained for classification.
As shown in the table, \method outperforms other SSL methods by up to $+13.04\%$ top-1 accuracy, possibly because it utilizes the knowledge from the FP network.
Interestingly, \method even outperforms past supervised ImageNet classification by binary networks \eg, XNOR-Net ($51.20\%$)~\cite{Rastegari2016XNORNetIC}.

\begin{table*}[t!]
    \centering
    \resizebox{0.7\linewidth}{!}{
        \begin{tabular}{cccccc}
        \toprule
 \multirow{2}{*}{Method}             & \multicolumn{4}{c}{Object-Centric} & \multicolumn{1}{c}{Scene-Centric} \\ \cmidrule(lr){2-5} \cmidrule(lr){6-6}
                & CIFAR10 & CIFAR100 & CUB-200-2011 & Birdsnap & Places205 \\ \midrule
        Supervised Pre.     &78.30&	57.82&				54.64&36.90	 & 46.38 \\ \midrule
        SWAV&   75.78&	56.78	&	36.11&	25.54 & 46.90       \\
        BYOL  &  76.68&	58.18	&	38.80&	27.11 & 44.62          \\ 
        Tuned MoCov2 &78.29  &57.56  & 33.79 & 23.37 & 44.90  \\
        S2-BNN & 82.70 & 61.90& 47.50 & 34.10& 46.58 \\ 
        \cellcolor{Orange!20}\method (Ours)  & \cellcolor{Orange!20}{\bf 84.60}&\cellcolor{Orange!20}{\bf61.99}	&\cellcolor{Orange!20}{\bf49.62}&\cellcolor{Orange!20}{\bf34.48} & \cellcolor{Orange!20}{\bf 47.22}      \\\bottomrule
        \\
        \end{tabular}
    }
    \vspace{-1.5em}
    \caption{Transfer learning (top-1) on either object-centric or scene-centric datasets after pretraining. CIFAR10, CIFAR100, CUB-200-2011, and Birdsnap are used as the object-centric datasets while Places205 is used as the scene-centric dataset. \method outperforms all other SSL baselines on the object-centric datasets and on Places205.}
    \vspace{-1em}
\label{tab:transfer}
\end{table*}

\vspace{-1em}\paragraph{Semi-Supervised Fine-Tuning.}
We now conduct semi-supervised fine-tuning (top-1 and top-5) and summarize the results in Table~\ref{tab:cls_semi}. 
We fine-tune the entire network on the labeled subset ($1\%$ or $10\%$) from ImageNet.
\method outperforms other SSL baselines by large margins across all metrics; at least $+3.67 \%$ top-1 accuracy and $+5.30 \%$ top-5 accuracy on the $1\%$ labels setting and $+3.98\%$ top-1 accuracy and $+4.41\%$ top-5 accuracy on the $10\%$ labels setting, respectively.
Interestingly, \method seems to outperform other SSL methods by larger amounts in this task than the linear evaluation, implying that \method may be more beneficial in tasks with limited supervision as also discussed by \cite{goyal2021self}.

\vspace{-1em}\paragraph{SVM Image Classification.}
We conduct SVM classification (mAP (\%)) and summarize results for both the few-shot and full-shot (`Full') settings on VOC07 in Table~\ref{tab:fewshot}.
For the few-shot results, the results are averaged over 5 runs. 
The number of shots $k$ is varied from $1$ to $96$.

For the few-shot setting, \method outperforms all other SSL methods by roughly $+1\%$ to $+10\%$ mAP depending the number of shots. 
Noticeably, \method performs very close to the supervised pretraining regardless of the number of shots.
This is consistent with the semi-supervised fine-tuning results; \method shows strong performance in tasks with limited supervision such as the few-shot classification~\cite{goyal2021self}.
In the full-shot setting, \method outperforms other SSL methods by up to $+8.33\%$ mAP and performs very similarly to supervised pretraining.
In both settings, representations learned with ImageNet by \method is still effective on a different dataset such as VOC07, potentially due to \method using a FP network to obtain targets that are generally useful \ie, low to mid level representations~\cite{zhao2021what}.

\begin{table*}[t!]
    \centering
    \resizebox{0.8\linewidth}{!}{
    \begin{tabular}{lccccc}
    \toprule
     Method             & \numcircledmod{1} Rand. Init. Cls. & \numcircledmod{2} Feat. Sim. Loss & \numcircledmod{3} Dyn. Bal. & \numcircledmod{4} Multi-Stage & Top-1 (\%)  \\ 
     \cmidrule(lr){1-5} \cmidrule(lr){6-6}
    ~\numcircledmod{1} &  {\color{ForestGreen} \cmark}&  {\color{red} \xmark}& {\color{red} \xmark} & {\color{red} \xmark} &57.33    \\ 
    ~\numcircledmod{1}+\numcircledmod{2} & {\color{ForestGreen} \cmark}& {\color{ForestGreen} \cmark}& {\color{red} \xmark} & {\color{red} \xmark} &   59.53                          \\ 
    ~\numcircledmod{1}+\numcircledmod{2}+\numcircledmod{3} & {\color{ForestGreen} \cmark}& {\color{ForestGreen} \cmark}& {\color{ForestGreen} \cmark} & {\color{red} \xmark} &    61.43                          \\
    ~\cellcolor{Orange!20}\numcircledmod{1}+\numcircledmod{2}+\numcircledmod{3}+\numcircledmod{4} (=\method)&\cellcolor{Orange!20}{\color{ForestGreen} \cmark} &\cellcolor{Orange!20}{\color{ForestGreen} \cmark} &\cellcolor{Orange!20}{\color{ForestGreen} \cmark} &\cellcolor{Orange!20}{\color{ForestGreen} \cmark} &\cellcolor{Orange!20}{\bf62.29} \\ \bottomrule
    \\
    \end{tabular}
    }
    \vspace{-1em}
    \caption{Ablation studies on the proposed components of \method using linear evaluation (top-1) on ImageNet. \numcircledmod{1} refers to using a randomly initialized classifier as targets. \numcircledmod{2} `Feat. Sim.' refers to feature similarity loss (Eq.~\ref{eq:aug}). \numcircledmod{3} `Dyn. Bal.' refers to using the dynamic balancing. \numcircledmod{4} refers to using the modified multi-stage training. Each step of improving \method contribute to a non-trivial performance gain as the evaluation is done with the ImageNet dataset. Also, using only \numcircledmod{1} already outperforms all other SSL baselines except S2-BNN~\cite{shen2021s2}.}
    \vspace{-0.5em}
\label{tab:ablation}
\end{table*}

\vspace{-1em}\paragraph{Transfer Learning.}
While \cite{shen2021s2} use VOC07, we use Places205 instead of VOC07 to show more diverse transfer scenarios as we already show transfer learning results to VOC07 using SVM in Table~\ref{tab:fewshot}.
Given that we use ImageNet (object-centric) for pretraining, to evaluate the transferability of learned representations across domains, we use two types of datasets for the transferability experiments, \ie, object-centric and scene-centric datasets. 
Specifically, we use CIFAR10, CIFAR100, CUB-200-2011 and Birdsnap as the `object-centric datasets', and use Places205 as the scene-centric dataset.
Once we pretrain the binary feature extractor with ImageNet, the feature extractor is frozen and only the attached classifier is trained on the target datasets.

As shown in the table, \method outperforms all SSL methods on the object-centric datasets with large margins in CIFAR10, CUB-200-2011 and Birdsnap.
It implies that the representations learned using \method transfers well across multiple object-centric datasets. 
For the scene-centric dataset (Places205), we observed that the transfer learning results for methods vary less.
It is quite expected since ImageNet is object-centric, thus transferring knowledge to a scene-centric dataset may suffer from domain gap and performance marginally differ across methods.

\subsection{Further Analyses}
\label{sec:further_analysis}
We further investigate our method using linear evaluation (top-1) on ImageNet for detailed analyses. 

\vspace{-1em}\paragraph{Training Time.}
\begin{wrapfigure}{r}{0.2\textwidth}
\centering
    \vspace{-1.6em}
    \includegraphics[width=0.2\textwidth]{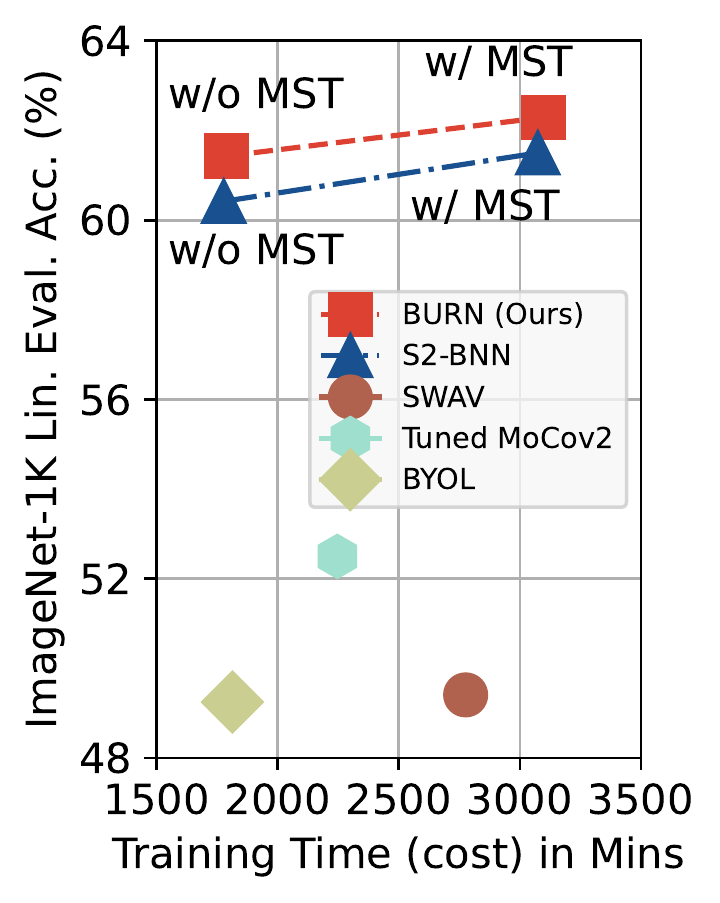}
    \vspace{-2.5em}
    \caption{Training time and linear evaluation accuracy.\vspace{-2em}} 
    \label{fig:running_time}
\end{wrapfigure}
A comparison of the training time \vs linear evaluation on ImageNet is shown in Fig.~\ref{fig:running_time}.
Without multi-stage (`MST'), BURN shows much higher accuracy than other baselines with similar or less training time. 
With MST, BURN shows higher accuracy than S2-BNN but comparable training time.

\vspace{-1em}\paragraph{Ablation Study.}
We ablate the model and summarize the results in Table~\ref{tab:ablation}.
We number each components, following the convention in Fig.~\ref{fig:setting}.
Our baseline method (\numcircledmod{1}) alone achieves $57.33\%$ top-accuracy, which is higher than all other SSL baselines except S2-BNN (see Table~\ref{tab:cls_semi}).
In addition, every component in \method contributes to non-trivial gains.
In particular, both the feature similarity loss and dynamic balancing provide noticeable improvements.
We believe the reason is that the addition of $\mathcal{L}_{FS}$ stabilizes the gradients (see Sec.~\ref{sec:augmented_loss}) and a dynamic balancing of $\mathcal{L}_{FS}$ captures the changing importance of $\mathcal{L}_{FS}$ stabilizing the gradients even more effectively, resulting in improved performance.
The modified multi-stage training also contributes to the accuracy by a non-trivial margin.

\vspace{-1em}\paragraph{Choice of Moving Targets for Binary Networks.}
\label{sec:mov_targ_choice}
Note that the downstream task performance of BYOL~\cite{grill2020bootstrap} and Tuned MoCov2~\cite{shen2021s2}, which use the EMA target and momentum encoder respectively, are worse than \method (see Sec.~\ref{sec:down_task}), which uses jointly trained classifier as moving targets.
In addition to the above quantitative comparison, we provide intuition for why the EMA target or the momentum encoder may be less effective choices as the moving target for binary networks.
The EMA target and the momentum encoder are both based on momentum or EMA updates of the target network and incur differences between the target and the backbone networks.
We conjecture that EMA updates may not be well-suited for binary networks because the differences of the target and backbone network from the EMA update could be amplified by the binarization process.

\begin{figure}[t!]
    \centering
    \resizebox{0.8\linewidth}{!}{
    \includegraphics[width=0.8\columnwidth]{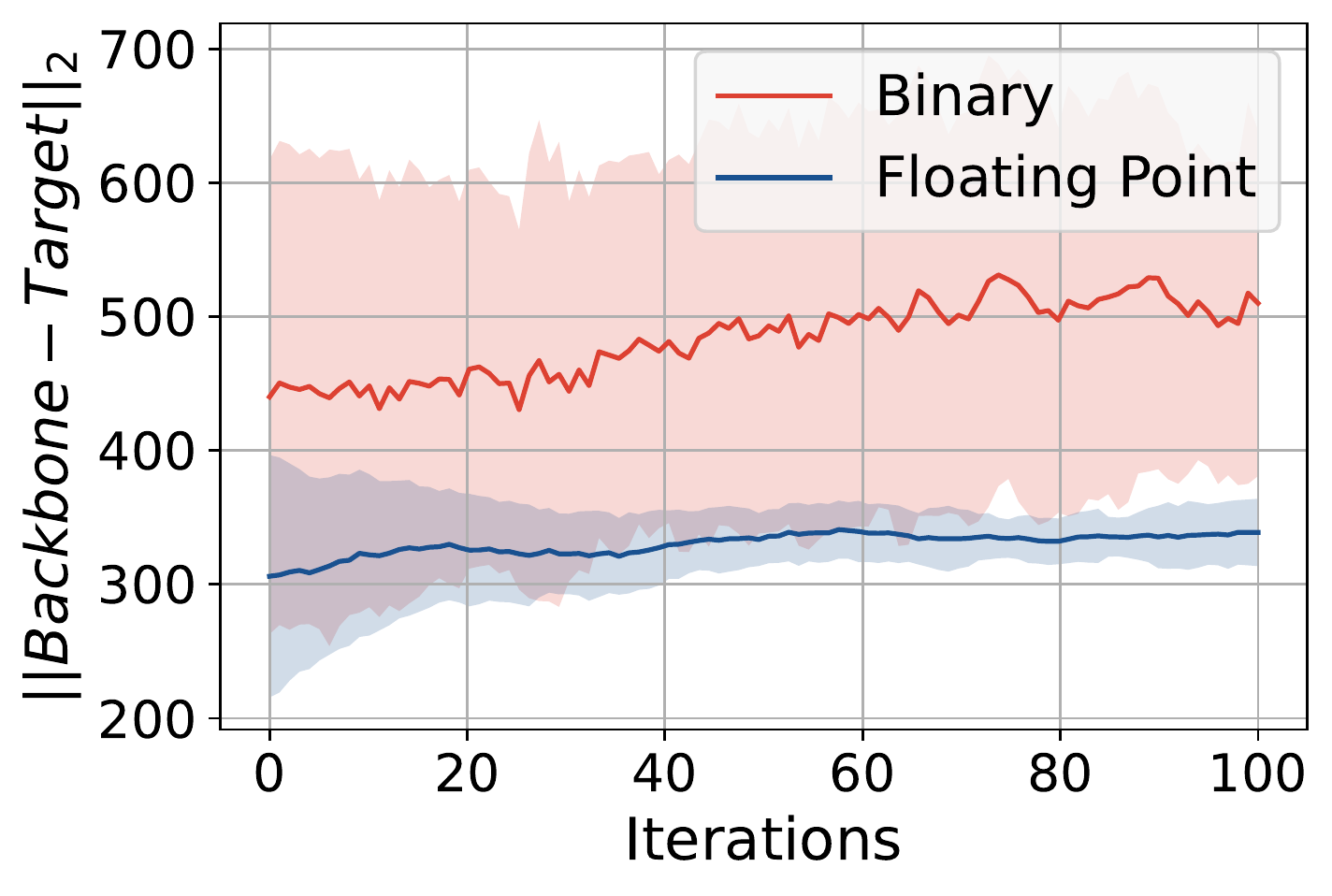}
    }
    \vspace{-1.2em}
    \caption{The L2 distance between the $\textit{Backbone}$ and $\textit{Target}$ vectors averaged over $10$ independent runs. The shaded areas denote the region of 1 std. The L2 distance is larger and has higher variance in the binary case, indicating that the the EMA target is more different from the backbone \ie, less effective as targets.}
    \label{fig:ema_bin}
    \vspace{-1.5em}
\end{figure}

To elaborate our intuition, we design an experiment to compare how different the backbone and EMA target networks are where the backbone and the EMA target are simplified to $100$-dimensional vectors denoted as $\textit{Backbone}$ and $\textit{Target}$, respectively.
We then simulate the training of the backbone and the EMA update of the target as:
\begin{equation}
\begin{split}
        \text{Backbone} & \leftarrow \text{Backbone} + \eta \cdot \Delta n, \\
    \text{Target} & \leftarrow \tau \cdot \text{Target} + (1-\tau) \cdot \text{Backbone},
    \end{split}
    \label{eq:toy_update}
\end{equation}
where $\textit{Backbone}$ and $\textit{Target}$ are initialized with the same random values, $\eta$ is the learning rate set as $4.8$~\cite{grill2020bootstrap}, $\Delta n$ is a random perturbation drawn from a standard normal distribution, $\mathcal{N}(0,\mathbf{1})$, to simulate the gradient update, and $\tau$ is the momentum used in the EMA update set as $0.99$.
We perform the update described in Eq.~\ref{eq:toy_update} for $100$ iterations. 

We compare the L2 distance between the $\textit{Backbone}$ and $\textit{Target}$ vectors in Fig.~\ref{fig:ema_bin}.
The L2 distance between the $\textit{Backbone}$ and $\textit{Target}$ is much larger with higher variance for the binary case, indicating that the EMA target may differ heavily from the backbone with binary networks.
Not only that, in the binary case, the L2 distance does not decrease over time.
These results are in-line with our intuition and empirical results that the EMA target could be more different for binary networks and hence be less effective.

\begin{table}[t!]
    \begin{minipage}{.4\linewidth}
        \captionsetup{width=.90\linewidth}
        \centering
        \resizebox{0.9\linewidth}{!}{
        \begin{tabular}{ccc}
        \toprule
         $\mathcal{L}_{FS}$   & Bounded          & Top-1 (\%)  \\ \midrule 
        $L_1$ & {\color{red} \xmark}&    51.46    \\ 
        $L_2$ &{\color{red} \xmark} &   50.28                          \\ 
        Cosine & {\color{ForestGreen} \cmark}&   {\bf62.29}\\
        \bottomrule
        \end{tabular}
        }
        \vspace{-1em}
        \caption{$L_1$, $L_2$, and the cosine distances are compared. The cosine distance is by far the best choice amongst the three, as is supported by our intuition that a bounded loss term would be better as the feature similarity loss.} 
        \vspace{-1em}
        \label{tab:aug_loss}
    \end{minipage}%
    \begin{minipage}{.6\linewidth}
        \captionsetup{width=.90\linewidth}
        \centering
        \resizebox{0.85\linewidth}{!}{
        \begin{tabular}{cc}
        \toprule
         $\lambda(t)$           & Top-1 (\%)  \\ \midrule 
        Constant             &    55.83    \\ 
        Heaviside Step: $H(-t+T_{max}/2)$  &   55.60                       \\ 
       Eq.~\ref{eq:dynamic_lambda}     &   {\bf62.29}                         \\
        \bottomrule
        \end{tabular}
        }
        \vspace{-1em}
        \caption{Comparison of dynamic balancing functions. Please refer to the supplement for a plot comparing the choices of $\lambda(t)$. The Eq.~\ref{eq:dynamic_lambda} (smooth annealing) is the best amongst the three choices. The constant function does not capture the dynamic nature of the balancing factor and the Heaviside step function disrupts the training midway due to the discontinuity.}
        \label{tab:dynamic_balancing}
    \end{minipage}
    \vspace{-1.5em}
\end{table}

\vspace{-1em}\paragraph{Choice of Feature Similarity Loss.} 
\label{sec:exp_choice_fs}
We further investigate the choices of the feature similarity loss ($\mathcal{L}_{FS}$ in Eq.~\ref{eq:final_obj}) in Table~\ref{tab:aug_loss}.
Besides the  cosine distance, \ie, $1-{\langle v_1, v_2 \rangle \over \|v_1\|_2 \cdot \|v_2\|_2}$, used in \method, we compare the $L_1$, \ie, $\|v_1-v_2\|_1$, and $L_2$, \ie, $\|v_1-v_2\|_2$, distances.
We believe that as both the $L_1$ and $L_2$ distances are not bounded, they may potentially cause problems such as gradient explosion, leading to the worse performance unlike the cosine distance.

\begin{figure}[t!]
    \centering
    \resizebox{0.96\linewidth}{!}{
    \includegraphics{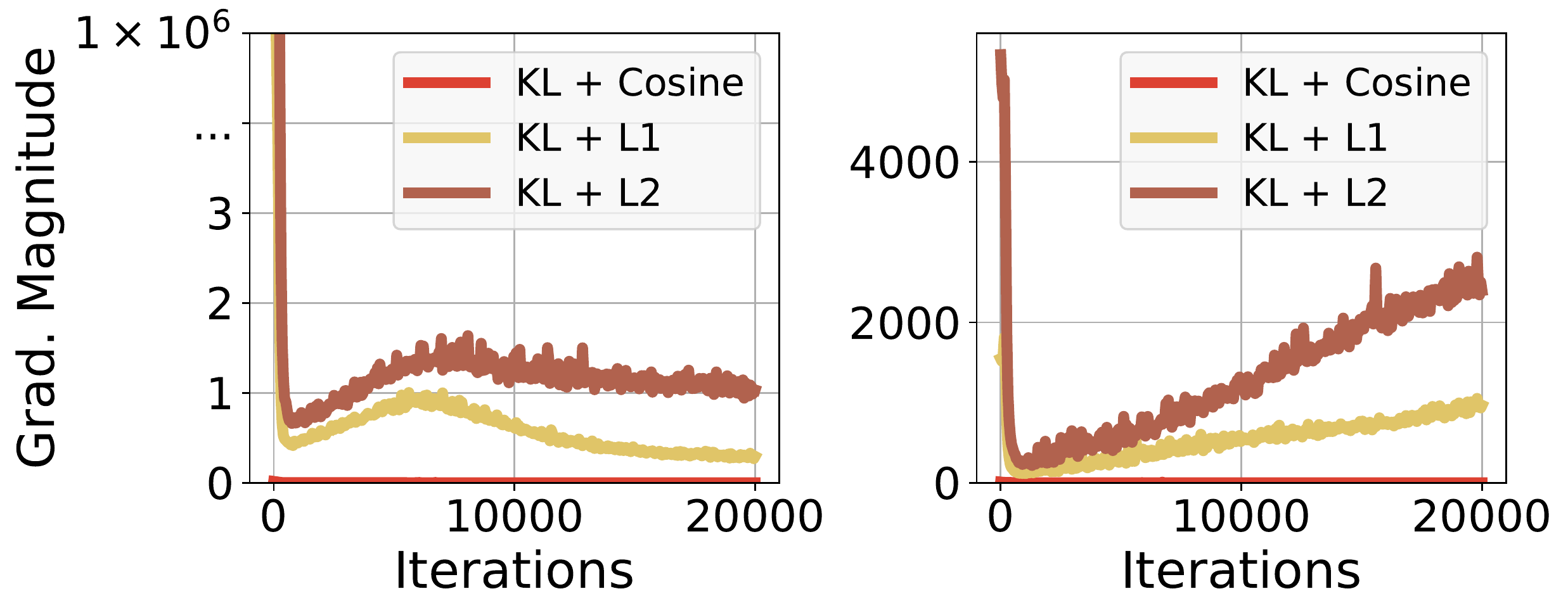}
    }\\
    {\footnotesize \hspace{5em}(a) Classifier\hspace{7em}(b) Feature extractor}
    \vspace{-0.5em}
    \caption{Gradient magnitude of (a) binary classifier and (b) the binary feature extractor during early training for various choices of $\mathcal{L}_{FS}$ such as the cosine, $L_1$, and $L_2$ distances. Both $L_1$ and $L_2$ distances show very high gradients at the beginning in the classifier, especially $L_2$. Moreover, $L_1$ and $L_2$ distances exhibit potential gradient explosions in the feature extractor. Cosine distance shows none of these trends that harm training efficacy.}
    \vspace{-1em}
    \label{fig:grad_ablation}
\end{figure}

Cosine distance outperforms $L_1$ and $L_2$ by large margins.
In Fig.~\ref{fig:grad_ablation}, we illustrate gradient magnitudes of both the classifier and feature extractor when using cosine, $L_1$, or $L_2$ distances as $\mathcal{L}_{FS}$.
$L_1$ and $L_2$ distances show very high gradients early on in the classifier, especially $L_2$ where the gradients start at $1\times10^6$. 
Even more importantly, $L_1$ and $L_2$ distances show signs of gradient explosions in the feature extractor, \ie, the gradients keep increasing as the iterations proceed, with $L_2$ exhibiting more severe trends.
In contrast, cosine distance exhibits small and subdued gradients for both the classifier and the feature extractor.

\vspace{-1em}\paragraph{Choice of the Dynamic Balancing Function.}
\label{sec:choice_of_lamdbda}
We finally investigate the choice of the dynamic balancing function.
Specifically, we consider (1) a constant function, $\lambda(t) = 0.7$, (2) a Heaviside step function that is shifted and horizontally reflected, $H(-t+T_{max}/2)$, and (3) a smooth cosine annealing function (Eq.~\ref{eq:dynamic_lambda}) used in our \method, and compare their respective accuracy in Table~\ref{tab:dynamic_balancing}.

The constant function fails to capture that the importance of $\mathcal{L}_{FS}$ may change as learning progresses, leading to poor results.
The Heaviside step function abruptly changes the balancing factor mid-training.
This disrupts the training and leads to poor performance.
In contrast, smooth annealing (Eq.~\ref{eq:dynamic_lambda}) captures the dynamic nature of the importance of the feature similarity loss while smoothly varying the balancing factor, resulting in the best performance among the options.

\section{Conclusion} 
\label{sec:conclusion}
We propose \method, an unsupervised representation learning framework for binary networks that jointly trains the FP classifier and the binary network.
We propose a feature similarity loss, dynamic loss balancing, and a modified multi-stage training to improve \method.
We conduct extensive empirical validations with five downstream tasks and seven datasets.
In all downstream tasks, \method consistently outperforms existing SSL baselines by large margins and sometimes even the supervised pretraining. 
We also provide detailed analysis on various aspects of \method.

\vspace{-1em}\paragraph{Limitations.}
One limitation of \method is that it requires a pretrained FP feature extractor. 
While using pretrained FP networks is a commonly used approach for binary networks~\cite{shen2021s2, liu2020reactnet, Martinez2020Training, Bulat2020BATSBA}, it would be more efficient to design a framework that does not require a pretrained FP model.

\vspace{-1em}\paragraph{Potential Negative Societal Impact.}
We aim to improve the representation power learned by binary networks, which would facilitate AI on edge devices with vision sensors.
Consequently, AI surveillance systems could become prevalent and may lead to the monitoring of mass populations for private information and personal attributes. 
Although the authors have \emph{no intention} to allow such consequences, such negative effects could manifest.
Efforts to prevent such effects would include gating the code and pretrained models behind security and ethical screenings. 

\vspace{0.5em}
\small{\noindent\textbf{Acknowledgement.} The authors thank Jung-Woo Ha for the valuable discussions. This work was partly supported by the National Research Foundation of Korea (NRF) grant funded by the Korea government (MSIT) (No.2022R1A2C4002300) and Institute for Information \& communications Technology Promotion (IITP) grants funded by the Korea government (MSIT) (No.2020-0-01361-003 and 2019-0-01842, Artificial Intelligence Graduate School Program (Yonsei University, GIST), and No.2021-0-02068 Artificial Intelligence Innovation Hub).}




{\small
\bibliographystyle{ieee_fullname}
\bibliography{egbib}
}

\newpage

\onecolumn
\appendix
\section{Details on Downstream Task Configurations and Implementation Details (Section~\ref{sec:exp})}
\label{sec:downstream_detail}
We present detailed explanations for each downstream task and the implementation.
For the experiments, we strictly follow the experimental protocols from relevant prior work~\cite{chen2020simple, chen2020big, goyal2019scaling, he2020momentum, xiao2021what} whenever possible.

\paragraph{Object Detection.} Following~\cite{he2020momentum}, we use the Faster R-CNN object detection framework. The Faster R-CNN framework is implemented using detectron2. We use Pascal VOC 2007 and Pascal VOC 2012 as the training dataset and evaluate on the Pascal VOC 2007 test set. We use the pretrained weights as the initial weights and fine-tune the entire detection framework. We use the exact same configuration file from~\cite{he2020momentum}.

\paragraph{Linear Evaluation.} Following~\cite{goyal2019scaling}, we attach a linear classifier (a single fully-connected layer followed by a softmax) on top of the frozen backbone network and train only the classifier for 100 epochs using SGD. The initial learning rate is set to 30 and multiplied by 0.1 at epoch 60 and 80. The momentum is set to 0.9 with no weight decay. The classifier is trained on the target datasets \ie, ImageNet and ImageNet100~\cite{Tian2020Contrastive}.

\paragraph{Semi-Supervised Fine-Tuning.} Following~\cite{chen2020simple, chen2020big, xiao2021what} we attach a linear classifier (a single fully-connected layer followed by a softmax) on top of the backbone network and fine-tune the backbone as well as the linear classifier using SGD for 20 epochs. 
Different initial learning rates are used for the backbone and the linear classifier. 
We select one from \{0.1, 0.01, 0.001\} for the backbone's initial learning rate. 
We multiply either \{1, 10, 100\} to the backbone's initial learning rate for the linear classifier initial learning rate.
We found the performance for different pretraining methods to vary considerably for the different learning rate configurations and hence we sweep all the 9 combinations of initial learning rates described above and use the best configuration for each method for fine-tuning.

The momentum is set to 0.9 with a weight decay of 0.0005 and the learning rates for the backbone and the classifier are multiplied by 0.2 at epochs 12 and 16.
For fine-tuning, only 1\% or 10\% of the labeled training images that are randomly sampled from the target datasets are used.
The entire validation set is used for evaluation.

\paragraph{SVM Image Classification.} Following~\cite{goyal2019scaling}, we first extract features from the backbone network and apply average pooling to match the feature vector dimension to be 4,096. 
Note that~\cite{goyal2019scaling} uses ResNet50 as a backbone and extracts features after each residual block to report the best accuracy. 
In contrast, we are based on ReActNet~\cite{liu2020reactnet} and extract features from the last layer as we found that to perform the best.

The feature vector dimension is 4,096 instead of 8,192 as in~\cite{goyal2019scaling} because of the backbone architecture difference.
With the extracted features, we use the LIBLINEAR~\cite{liblinear08} package to train linear SVMs.
For the `full-shot' classification, we use the `trainval' split of VOC07 dataset for training and evaluate on the `test' split of VOC07 dataset.
We report the mAP for the full-shot classification.
For the few-shot classification, we use the `trainval' split of VOC07 dataset in the few-shot setting for training and evaluate on the `test' split of VOC07 dataset.
The number of shots $k$ (per class) varies from 1 to 96. 
We report average of mAP over five independent samples of the training data along with the standard deviation for the few-shot classification.

\paragraph{Transfer Learning.} We perform linear evaluation on various target datasets.
For object-centric datasets (\eg, CIFAR10, CIFAR100, CUB-200-2011, Birdsnap), following~\cite{goyal2019scaling}, we attach a linear classifier (a single fully-connected layer followed by a softmax) on top of the frozen backbone network and train only the classifier for 100 epochs using SGD. The initial learning rate is set to 30 and multiplied by 0.1 at epoch 60 and 80. The momentum is set to 0.9 with no weight decay. The classifier is trained on the target datasets.
For scene-centric datasets (\eg, Places205), following~\cite{goyal2019scaling}, we attach a linear classifier on top of the frozen backbone network and train only the classifier. We train the classifier for 28 epochs using SGD. The learning rate is set to 0.01 initially and is multiplied by 0.1 at every 7 epochs. The momentum is set to 0.9 with weight decay of 0.00001. Note that different hyper-parameters are used from the object-centric datasets, following~\cite{goyal2019scaling}. The classifier is trained on the target datasets.

\paragraph{Implementation Details.}
For our binary network backbone, we use authors' implementation of the ReActNet-A~\cite{liu2020reactnet} in all of our experiments.
For pretraining, we use the LARS optimizer~\cite{you2017large} with a batch size of $2,048$ for $200$ epochs on ImageNet with the learning rate set as $0.3$.
For the FP network, we use a ResNet50 pretrained using MoCov2~\cite{he2020momentum} on ImageNet for $800$ epochs.
We use $\lambda_{0} = 0.9$ and $\lambda_{T_{max}} = 0.7$ but other values performed similarly, given $\lambda_{T_{max}}$ was sufficiently smaller than $\lambda_{0}$.
The implementation is largely based on an earlier version of \cite{mmselfsup2021}.

\section{Additional Comparisons to InfoMin and SimCLRv2 (Section~\ref{sec:exp})}
\label{sec:add_comp_to_infomin_simclr}
We present additional comparisons to the InfoMin~\cite{tian2020makes} and SimCLRv2~\cite{chen2020big} with the ReActNet backbone to BURN (Ours) in Tables~\ref{tab:cls_semi_supple} and \ref{tab:is_fewshot}.
Note that we use ImageNet100~\cite{Tian2020Contrastive} to pretrain the models instead of ImageNet for faster experiments.
We use a ResNet50 trained for 600 epochs on ImageNet100 using BYOL as our pretrained FP network for \method.
We found that most methods perform poorly on the object detection and the transfer learning tasks, likely due to the the insufficient amount of pretraining data (\eg, ImageNet100).
Thus, we report results on linear evaluation, semi-supervised fine-tuning, and SVM image classification tasks but not on the object detection and transfer learning tasks.

Note that InfoMin and SimCLRv2 do not perform well in all the tested downstream tasks compared to \method, with InfoMin scoring much lower across the board.
We believe one reason for this is that the binary network lacks sufficient capacity to solve the instance discrimination task used in InfoMin and SimCLRv2; a similar problem was also found in lower capacity FP networks~\cite{compress, fang2021seed}.

\paragraph{Linear Evaluation.}
As shown in Table~\ref{tab:cls_semi_supple}, \method outperforms both InfoMin and SimCLRv2 by large margins, \textit{i.e.}, over 30\% for InfoMin and 14\% for SimCLRv2 in the top-1 accuracy.
Not only that, it even outperforms the supervised pretraining. 
InfoMin and SimCLRv2 perform poorly, with InfoMin scoring particularly low (see Section~\ref{sec:add_inves_infomin} for a detailed discussion).

\paragraph{Semi-Supervised Fine-tuning.} As shown in Table~\ref{tab:cls_semi_supple}, \method outperforms InfoMin and SimCLRv2 by large margins (\textit{e.g.}, over 40\% for InfoMin and almost 20\% for SimCLRv2 in top-1 accuracy in the 1\% label setting).
Additionally, \method performs similar to the supervised pretraining whereas other compared SSL methods do not.
Interestingly, SimCLRv2 performs much better than InfoMin, possibly because of the deeper projection layer, which was proposed in~\cite{chen2020big} to improve semi-supervised fine-tuning performance.
In contrast, InfoMin performs poorly compared to other methods.

\begin{table}[t!]
    \centering
    \resizebox{0.75\linewidth}{!}{
\begin{tabular}{ccccccc}
\toprule
\multirow{3.5}{*}{Pretrain on} & \multirow{3.5}{*}{Method}  & \multirow{2}{*}{Linear Eval.} & \multicolumn{4}{c}{Semi-Supervised Fine-tuning}\\
& & & \multicolumn{2}{c}{1\% Labels} & \multicolumn{2}{c}{10\% Labels} \\ 
 \cmidrule(lr){3-3} \cmidrule(lr){4-5}\cmidrule(lr){6-7}
& & Top-1 (\%) & Top-1 (\%) & Top-5 (\%) & Top-1 (\%) & Top-5 (\%) \\ 
 \cmidrule(lr){1-2}\cmidrule(lr){3-3}\cmidrule(lr){4-5}\cmidrule(lr){6-7}
\multirowcell{4}{ImgNet100}  &  Supervised Pre. &76.54   &63.10&86.24&				75.40&	92.16 \\
 \cmidrule(lr){2-2}\cmidrule(lr){3-3} \cmidrule(lr){4-5}\cmidrule(lr){6-7}
        &InfoMin~\cite{tian2020makes}& 45.38 & 21.68	&46.74&				32.06&	59.82           \\ 
        &SimCLRv2~\cite{chen2020big} & 61.40&    43.78	&72.28&				60.06&	84.9           \\ 
 &   \cellcolor{Orange!20}\method (Ours)  & \cellcolor{Orange!20}{\bf77.02}&\cellcolor{Orange!20}{\bf 62.60}&\cellcolor{Orange!20}{\bf84.75}	&\cellcolor{Orange!20}{\bf74.00}&\cellcolor{Orange!20}{\bf91.38}       \\
\bottomrule
\end{tabular}
}
\vspace{-0.5em}
\caption{Linear evaluation (top-1) and semi-supervised fine-tuning (1\% labels or 10\% labels) on ImageNet100 after pretraining. \method outperforms all other SSL methods by large margins across for both the linear evaluation and semi-supervised  fine-tuning. The best result among SSL methods is shown in {\bf bold}.}
\vspace{-0.5em}
\label{tab:cls_semi_supple}
\end{table}

\begin{table}[t!]
\captionsetup{width=0.99\linewidth}
    \centering
    \resizebox{0.99\linewidth}{!}{
\begin{tabular}{ccccccccccc}
\toprule
Pretrain on & Method          & k = 1& k = 2 & k = 4 & k = 8 & k = 16 & k = 32 & k = 64 & k = 96 & Full\\ \midrule
\multirowcell{4}{ImgNet100}&Supervised Pre.     &	22.18 $\pm$ 1.31&	28.77 $\pm$ 1.97&	36.59 $\pm$ 1.61&	43.67 $\pm$ 0.93	&50.61 $\pm$ 0.62&	55.75 $\pm$ 0.43&	59.39$\pm$ 0.20&	60.88$\pm$ 0.41 & 64.77\\ \cmidrule{2-11}
&InfoMin & 14.12$\pm$ 0.23&	17.07$\pm$0.93&	20.76$\pm$ 0.91&	24.75 $\pm$ 0.27&	29.9 $\pm$0.73&	35.12$\pm$0.52&	39.2 $\pm$0.31&	41.90$\pm$ 0.22 & 47.32\\
&SimCLRv2  & 17.97 $\pm$ 0.56&	22.87 $\pm$2.0&	30.48 $\pm$ 1.02&	34.98 $\pm$1.58&	42.9 $\pm$1.03&	48.81 $\pm$0.67&	53.87 $\pm$0.48&	56.21 $\pm$0.25 & 61.36\\ 
&\cellcolor{Orange!20}\method (Ours)&\cellcolor{Orange!20} {\bf25.47 $\pm$ 1.46}&\cellcolor{Orange!20}{\bf30.68 $\pm$ 1.96}&\cellcolor{Orange!20}{\bf39.01 $\pm$ 1.28}&\cellcolor{Orange!20}{\bf46.13 $\pm$ 1.54}&\cellcolor{Orange!20}{\bf52.90 $\pm$ 0.81}&\cellcolor{Orange!20}{\bf58.29 $\pm$0.40}&\cellcolor{Orange!20}{\bf63.21 $\pm$ 0.30}&\cellcolor{Orange!20}{\bf64.96 $\pm$ 0.09} & \cellcolor{Orange!20}{\bf69.56}\\ \bottomrule
\end{tabular}
}
\caption{SVM image classification using SVM (mAP) on the VOC07 dataset after pretraining on ImageNet100 is shown. The number of shots ($k$) is varied from 1 to 96. We report average mAP over 5 runs with the standard deviation. BURN outperforms all methods including the supervised pretaining. The best result among SSL methods for each shot is shown in {\bf bold}.}
\label{tab:is_fewshot}
\end{table}

\paragraph{SVM Image Classification.} We show the SVM image classification results for both the few-shot and full-shot (`Full') in  Table~\ref{tab:is_fewshot}.
For the full-shot setting, \method outperforms both InfoMin and SimCLRv2 by large margins, \textit{i.e.}, over 21\% for InfoMin and 7\% for SimCLRv2 on mAP.

For the few-shot results, \method outperforms both InfoMin and SimCLRv2 by large margins across all number of shots including the full-shot.
Again, InfoMin performs worse than the other SSL methods as well, achieving mAPs that are well below the mAPs of \method or SimCLRv2 in all numbers of shots ($k$) and the full-shot setting (please refer to  Section~\ref{sec:add_inves_infomin} for discussion on the reason for InfoMin's poor performance).

\begin{figure}
    \centering
    \resizebox{0.5\linewidth}{!}{
    \includegraphics[width=0.8\columnwidth]{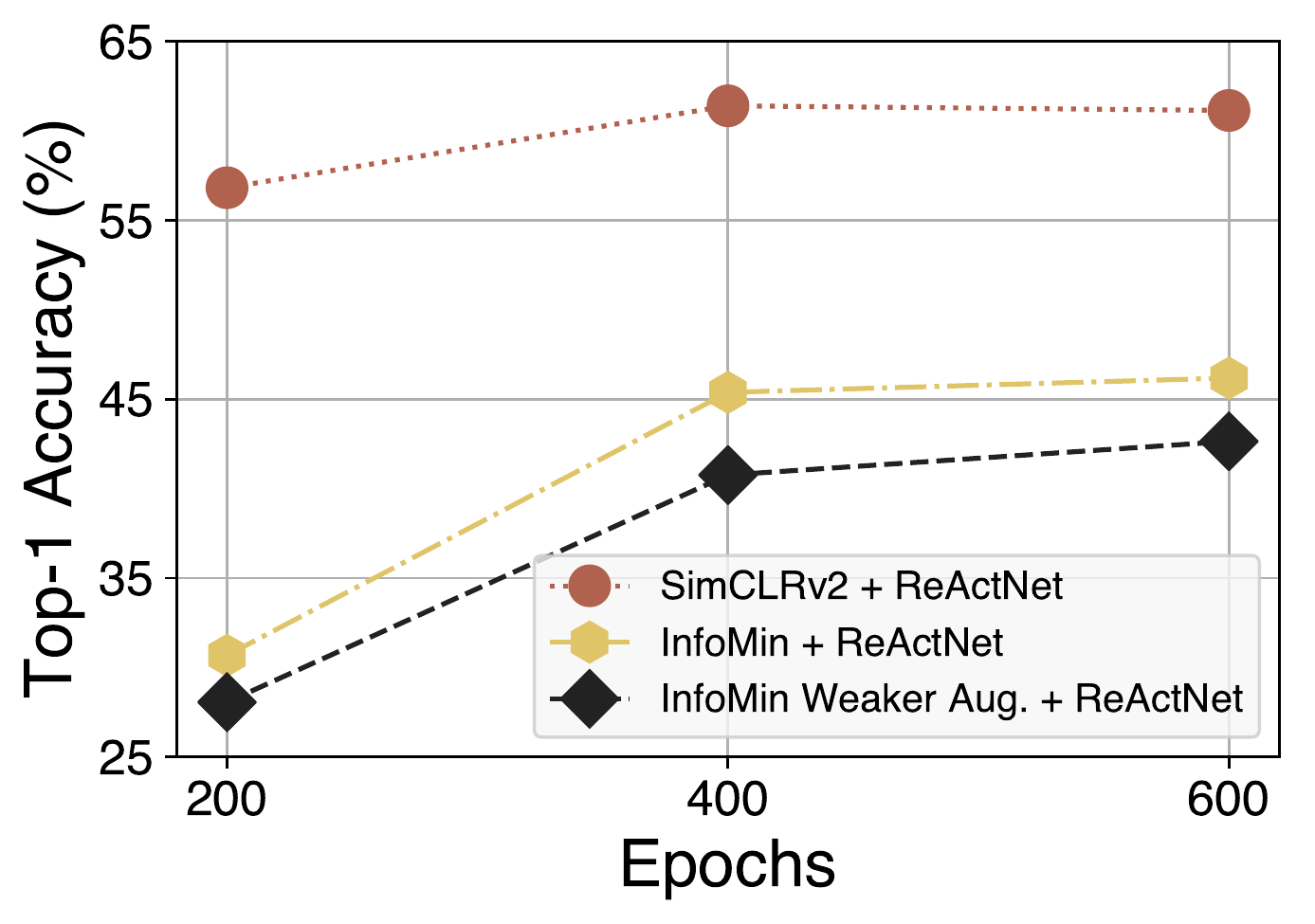}
    }
    \captionof{figure}{Comparison of linear evaluation performance on the ImageNet100~\cite{Tian2020Contrastive} dataset for InfoMin with a weaker augmentation strategy using ReActNet as the backbone. Weaker augmentation further degrades the performance of InfoMin.}
    \label{fig:infomin_aug}
\end{figure}

\subsection{Further Investigation on Performance Degradation of InfoMin (Follow-up of Section~\ref{sec:add_comp_to_infomin_simclr})}
\label{sec:add_inves_infomin}

\begin{figure}[t!]
    \centering
    \resizebox{0.8\linewidth}{!}{
    \includegraphics[width=0.99\linewidth]{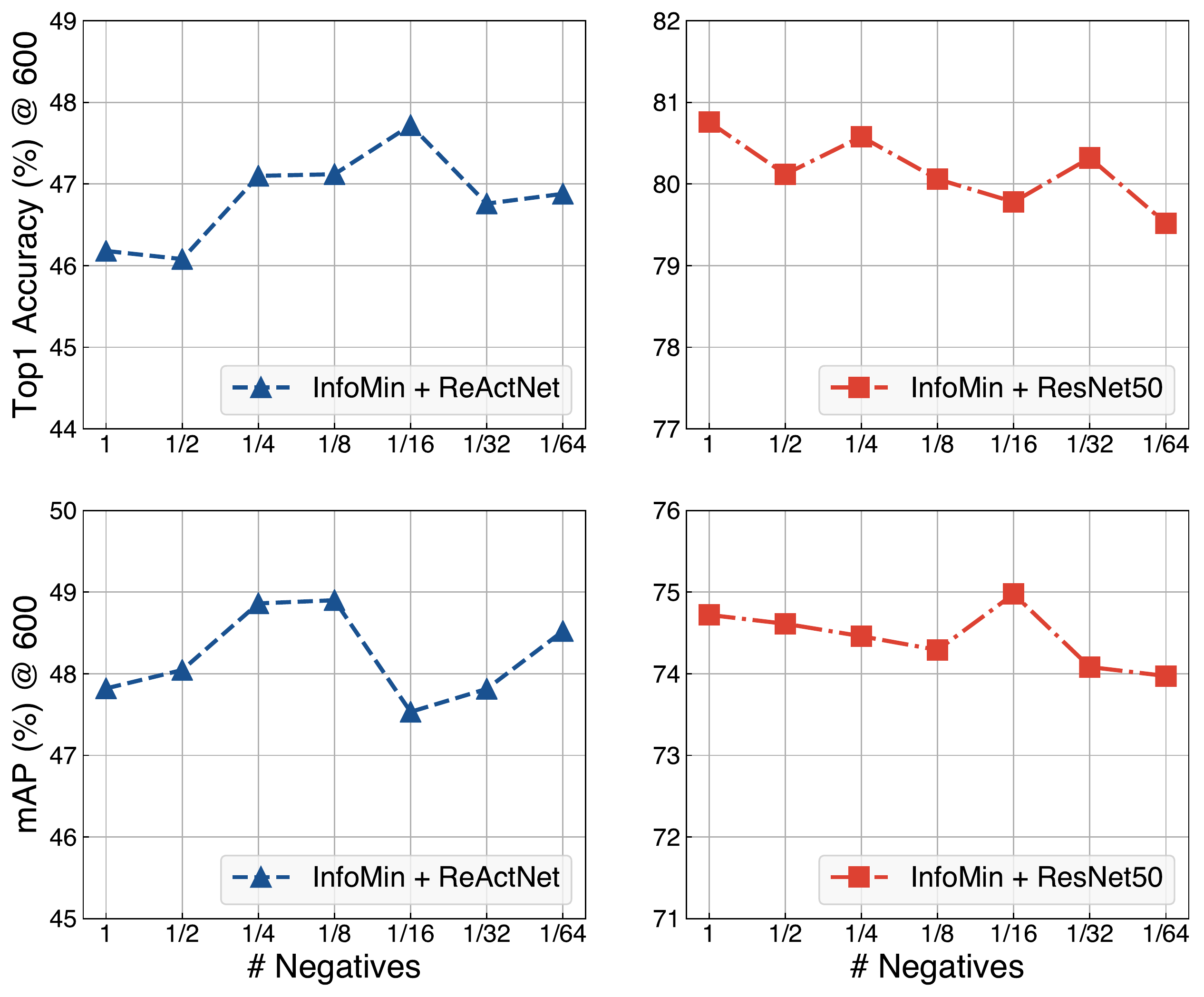}
    }
    \caption{Analysis of the number of negatives (\# Negatives) used in InfoMin with a binary network (ReActNet) or a FP network (ResNet50) on linear evaluation (top-1 (\%) and SVM classification(mAP (\%)) on the ImageNet100~\cite{Tian2020Contrastive}] and VOC07 datasets at epoch 600. Decreasing the number of negatives is helpful for binary networks while not being beneficial for FP networks.}
    \label{fig:infomin_negative}
\end{figure}

InfoMin with ReActNet as the backbone performs noticeably poorly as the results shown in Section~\ref{sec:add_comp_to_infomin_simclr}.
We first hypothesized in Section~\ref{sec:add_comp_to_infomin_simclr} that the performance degradation is due to the contrastive pretext task being too difficult for binary networks.
However, that does not fully explain the additional performance discrepancy between InfoMin and Simclr2 as they both use the contrastive pretext task.
Thus, we present further empirical analyses to investigate the performance degradation of InfoMin.

Note that the main differences between InfoMin and SimCLRv2 are 1) stronger augmentations and 2) the momentum encoder for supplying large amount of negative samples.

\paragraph{Effect of strong data augmentation.}
We first investigate the effects of using the `stronger augmentations' used in InfoMin as they can distort the images excessively.
The linear evaluation results (top-1 accuracy) are shown at epochs 200, 400 and 600 in Figure~\ref{fig:infomin_aug} for InfoMin with a weakened augmentation strategy, \textit{e.g.}, only using random cropping, color distortion and Gaussian blur, following~\cite{chen2020simple, chen2020big, xiao2021what}, instead of the stronger augmentation used in InfoMin.
SimCLRv2 is also compared as a reference.
As shown in the figure, weakening the augmentation strategy further degrades the performance of InfoMin at epochs 200, 400, and 600.
This implies that the stronger augmentation strategy used by InfoMin may not be the main cause of InfoMin's sudden performance drop.
Hence, the comparison motivates us to further investigate the use of the momentum encoder with binary networks.


\begin{figure}[t!]
    \centering
    \resizebox{1.0\linewidth}{!}{
    \includegraphics{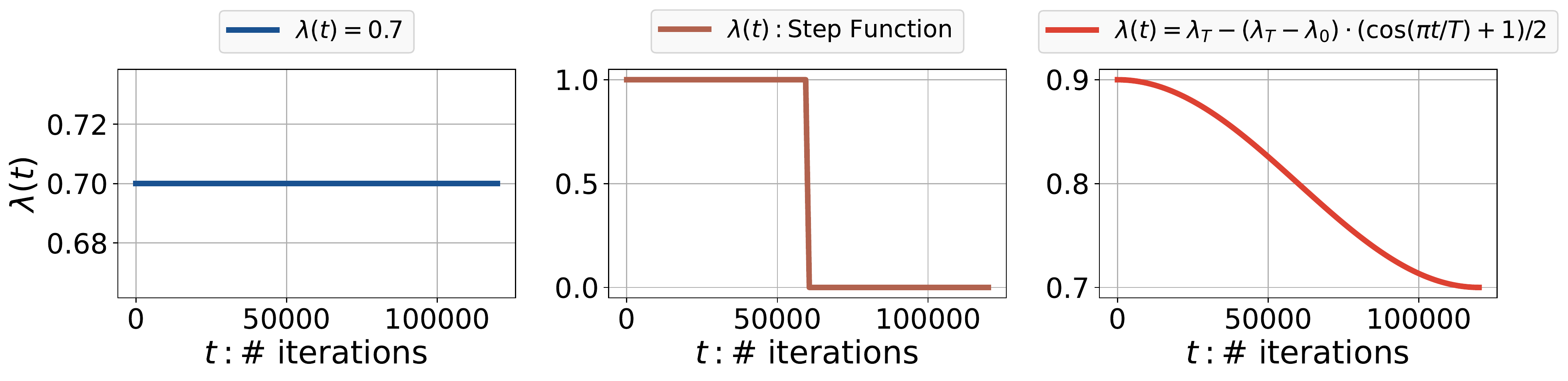}
    }
    {\footnotesize \hspace{10em}(a) Constant Function \hspace{8em}(b) Heaviside Step Function \hspace{8em}(c) Smooth Annealing Function}
    \caption{Plots for the various $\lambda(t)$ compared in Table~\ref{tab:dynamic_balancing} are shown.}
    \vspace{-1em}
    \label{fig:lambdas}
\end{figure}

\paragraph{Usage of momentum encoder.}
We then suspect that the use of momentum encoder to supply a large amount of negatives may not be beneficial in the binary case, as is also discussed in Sec.~\ref{sec:exp_choice_fs} when we investigated various choices for a moving target network.
Thus, using a momentum encoder may make the features of the negative samples too different to be helpful when it is used with binary networks, especially when the number of negative samples is large (65,536).
To empirically validate this, we vary the number of negative samples used in InfoMin to see the performance trend when either ReActNet (binary) or ResNet50 (FP) is used as the backbone in Fig.~\ref{fig:infomin_negative}.

As shown in the figures, InfoMin with ReActNet performs the best with a substantially smaller \# negatives ($1/16$) while with ResNet50 it performs worse with less \# negatives.
For further analysis of the effect made by the reduced number of negative samples, we evaluate the performance of SVM classification on the VOC07 dataset in the second row of Fig.~\ref{fig:infomin_negative}.
We observe similar trends in the SVM classification results.
While the above comparisons do not fully explain the InfoMin's performance drop, it is still interesting to see that using a momentum encoder to supply a large number of negatives may not be as helpful for binary networks.

\section{Plots of Various $\lambda(t)$ (Table~\ref{tab:dynamic_balancing} in Section~\ref{sec:exp_choice_fs})}
\label{sec:dynamic_lambda_plot}

In Fig.~\ref{fig:lambdas}, we plot (1) the constant function, (2) the Heaviside step function that is shifted and horizontally reflected, \ie,  $H(-t + T{\text{max}}/2)$ with $T_\text{max}$ being the maximum iterations, and (3) the smooth annealing function (Eq.~\ref{eq:dynamic_lambda}) to visualize the various choices of $\lambda(t)$ compared in Table~\ref{tab:dynamic_balancing}.

\end{document}